\documentclass{article}
\usepackage{spconf,amsmath,graphicx}

\usepackage{booktabs}
\usepackage{enumitem}
\usepackage{amsmath}
\setlist{nosep, leftmargin=14pt}

\usepackage{mwe} 
\usepackage{comment}

\usepackage{url}
\usepackage[hidelinks]{hyperref}
\usepackage{amsmath,amsthm,amssymb,amsbsy}
\usepackage{paralist}
\usepackage{xcolor}
\usepackage{color}
\usepackage{graphicx}
\graphicspath{{./figs/}}
\usepackage{algorithm}
\usepackage{algorithmic}
\usepackage{comment}

\usepackage{fancyhdr}
\usepackage{cite}
\usepackage{cleveref}

\newtheorem{theorem}{Theorem}

\theoremstyle{remark}

\makeatletter
\DeclareRobustCommand\onedot{\futurelet\@let@token\@onedot}
\def\@onedot{\ifx\@let@token.\else.\null\fi\xspace}

\def\etc{\emph{etc}\onedot} 
\def\wrt{w.r.t\onedot} 

\newcommand{\R}{\mathbb{R}}

\newcommand{\e}{\begin{equation}}
\newcommand{\ee}{\end{equation}}
\newcommand{\en}{\begin{equation*}}
\newcommand{\een}{\end{equation*}}
\newcommand{\eqn}{\begin{eqnarray}}
\newcommand{\eeqn}{\end{eqnarray}}
\newcommand{\bmat}{\begin{bmatrix}}
\newcommand{\emat}{\end{bmatrix}}

\newcommand{\ald}{\begin{aligned}}
\newcommand{\aldn}{\end{aligned}}

\newcommand{\vct}[1]{\boldsymbol{#1}}
\newcommand{\mtx}[1]{\boldsymbol{#1}}

\newcommand{\optr}[1]{\operatorname{\textbf{tr}}\left(#1\right)}
\newcommand{\optrb}[1]{\operatorname{\textbf{tr}}\left\{#1\right\}}

\newcommand{\vf}{\vct{f}}
\newcommand{\vg}{\vct{g}}

\newcommand{\vp}{\vct{p}}
\newcommand{\vq}{\vct{q}}

\newcommand{\vbeta}{\vct{\beta}}

\newcommand{\vx}{\vct{x}}

\newcommand{\mD}{\mtx{D}}

\newcommand{\mF}{\mtx{F}}
\newcommand{\mG}{\mtx{G}}

\newcommand{\mJ}{\mtx{J}}
\newcommand{\mK}{\mtx{K}}
\newcommand{\mL}{\mtx{L}}

\newcommand{\mP}{\mtx{P}}

\newcommand{\mW}{\mtx{W}}
\newcommand{\mX}{\mtx{X}}

\newcommand{\mZ}{\mtx{Z}}

\newcommand{\mPhi}{\mtx{\Phi}}

\graphicspath{{./figs/}}

\newlength{\imgwidth}
\newboolean{twoColVersion}
\setboolean{twoColVersion}{false}
\newcommand{\twoCol}[2]{\ifthenelse{\boolean{twoColVersion}} {#1} {#2} }

\usepackage[framemethod=tikz]{mdframed}

\title{Adaptive Weighted Multiview Kernel Matrix Factorization with its application in Alzheimer's Disease Analysis --- A clustering Perspective}
\name{Kai Liu\thanks{kail@clemson.edu, yaruic@clemson.edu}, Yarui Cao\thanks{Data used in preparation of this article were obtained from
		the Alzheimer’s Disease Neuroimaging Initiative (ADNI) database (adni.loni.usc.edu).}}  
\address{Author Affiliation(s)}

\address{Clemson University\\
	Computer Science Division\\
	Clemson, SC, USA}
\begin{document}
%
\maketitle
\begin{abstract}
Recent technology and equipment advancements provide with us opportunities to better analyze Alzheimer's disease (AD), where we could collect and employ the data from different image and genetic modalities that may potentially enhance the predictive performance. To perform better clustering in AD analysis, in this paper we propose a novel model to leverage data from all different modalities/views, which can learn the weights of each view adaptively. Different from previous vanilla Non-negative Matrix Factorization which assumes data is linearly separable,  we propose a simple yet efficient method based on kernel matrix factorization, which is not only able to deal with non-linear data structure but also can achieve better prediction accuracy. Experimental results on ADNI dataset demonstrate the effectiveness of our proposed method, which indicate promising prospects of kernel application in AD analysis.
\end{abstract}
\section{Introduction}
\label{sec:intro}
Alzheimer’s disease (AD) is a chronic neurodegenerative disease related with part of brain that controls memory, thought and language. It often happens in the elderly, beginning with mild memory loss and possibly leading to other cognitive functions loss, and getting worse dramatically. The whole progress can be divided into three diagnostic groups: AD, mild cognitive impairment (MCI), and health control (HC). 

\begin{figure}
	\centering
	\includegraphics[width=1\linewidth, height=5.5cm]{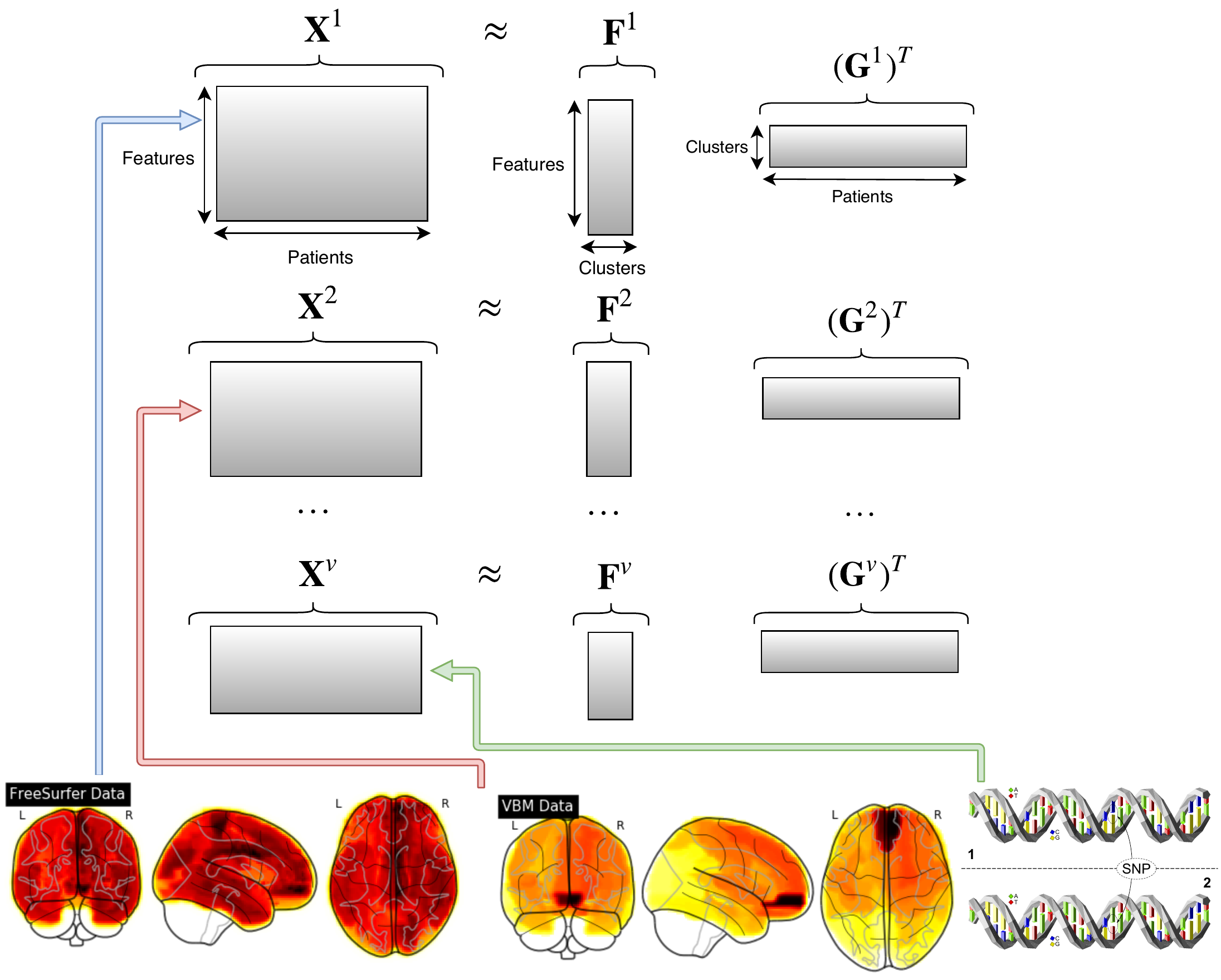}
	\caption{Multi-view Clustering via Matrix Factorization.}
	\label{fig:multiview}
\end{figure}

With the development of imaging genetics, exploring genome-wide array data and multimodal brain imaging data may help researchers deepen the understanding of AD, facilitate the accuracy of early detection and diagnosis, and improve its treatment. For example, ~\cite{nazarian2019genome} found significant association of certain single-nucleotide polymorphisms (SNPs) with AD. Magnetic resonance imaging (MRI) is an important medical neuroimaging technique to acquire regional imaging biomarkers, such as voxel-based morphometry (VBM) ~\cite{wang2015voxel} features to investigate focal structural abnormalities in grey matter (GM). Fluorodeoxyglucose positron emission tomography (FDG-PET) ~\cite{marcus2014brain} can distinguish AD from other causes of dementia with characteristic patterns of glucose metabolism. Therefore, multi-view analysis has attracted more attentions in recent years, and among those Non-negative matrix factorization (NMF) based approach is a promising one with strong interpretability which attracts more and more attention~\cite{liu2013multi,liu2018multiple,zong2017multi,liu2019study}.
However, existing multi-view analysis treat each view to be equally important, which contradicts the fact that some view may play more critical role than others. Therefore, learning the weights of different views adaptively becomes an interesting topic of importance.  On the other hand, vanilla NMF's success relies heavily on the assumption that data lies in a well-separable space with linear structure, but in practice, its accuracy is rather low, which indicates the necessity of developing new method which can deal with non-linearity data structure.

To address the problems aforementioned, in this paper, we propose a new method with efficient updating algorithm. Our contribution is twofold: first, the new framework can learn each view's weight and therefore conduct view selection; second, by introducing kernel mapping, it can handle both linear and non-linear cases. Extensive experiments on the Alzheimer’s Disease Neuroimaging Initiative (ADNI) dataset demonstrate its effectiveness with higher clustering results, which sheds a light on the prospect of Kernel application in other AD analysis tasks such as regression, classification, \etc 

\section{Related Works}

The framework in this paper originates from NMF, which is widely used in clustering as it can learn features and membership indicators in a very natural and interpretable way. Thus, we first give a brief overview of it along with its extensions. 

\subsection{Single View NMF with Graph Regularization}
Assume we have input data $\mX = [\vx_1, \vx_2, \cdots, \vx_n] \in \R^{m\times n}$ and the task is to cluster $\mX$ into $k$ clusters. Vanilla NMF assumes that each data point $\vx$ can be  represented as a linear combination of features $\vf$ (columns of $\mF$): $\vx_k \approx \sum_i \mF_i\mG_{ik}=\mF \vg_k$. Therefore, the  
 objective can be formulated as: 
\begin{equation}
	\label{graph_reg}
	\min \|\mX-\mF\mG \|_F^2 + \theta\optr{\mG\mL\mG^T} \quad s.t. \quad \mF,\mG \geq 0,
\end{equation}
where $\|\cdot\|_F^2$ denotes the squared Frobenius norm with $\|\mZ\|_F^2=\sum_i\sum_j z_{ij}^2$, and the second term `graph regularization' ~\cite{cai2010graph} was introduced to promote clustering accuracy. Each column of $\mF\in \R^{m\times k}$ represents a centroid of $k$ clusters, while each column of $\mG\in \R^{k\times n}$ can be taken as the probabilities of data $\vx$ belonging to each cluster. 
$\mL := \mD-\mW$ denotes the Laplace matrix, where $\mD$ is a
diagonal matrix whose entries are column (or row, since W
is symmetric) sums of $\mW$: $\mD_{ii} = \sum_j \mW_{ij}$, with $\mW\in\R^{n\times n}$ being the similarity matrix of data samples.\footnote{In our paper, we set $\mW_{ij}^{(a)} = exp(-\frac{\|\vx_i^{(a)} - \vx_j^{(a)} \|^2}{2\sigma^2})$.}
%


\subsection{Multiview Clustering via NMF}
Assume we have $v$ different views of input data $\{\mX^{(1)}, \mX^{(2)}, \\ \cdots, \mX^{(v)}\}$, where $\mX^{(i)}$ indicates the $i$-th view. For each view, we can factorize it as $\mX^{(i)} = \mF^{(i)} \mG^{(i)}$, where $\mF^{(i)} \in \R^{m\times k}$ contains the centroids of $k$ clusters, and $\mG^{(i)}\in \R^{k\times n}$ indicates the corresponding coefficient (probabilities matrix) for each view. 
%
As clustering results may vary with different views, a consensus clustering indicator matrix $\mG^*$ was introduced to promote the consistency across views. Moreover, similar to single view NMF, graph regularization term is added for each view and the objective can be formulated as:
\begin{equation}\label{multiview_reg}
\begin{aligned}
     \min_{\mF, \mG, \mG^*} &\sum_{a=1}^v\{\|\mX^{(a)}-\mF^{(a)}\mG^{(a)} \|_F^2 +  \lambda_a \|\mG^{(a)}-\mG^*\|_F^2\\
    + \ &\theta_a \ \optr{\mG^{(a)}\textbf{L}^{(a)}(\mG^{(a)})^T}\} \quad
    s.t. \ \ \mF^{(a)}, \mG^{(a)}, \mG^* \geq 0.
\end{aligned}
\end{equation}
where $\lambda, \theta$ are regularization parameters needs tuning. $\mG^{(a)}$ indicates clustering results from view $a$ and final consensus clustering will be obtained from $\mG^*$.

\section{Our Methodology}
Following the multi-view NMF clustering above, we applied it to Alzheimer's Disease dataset and found the clustering performance is not that high. Similar result can be found from \cite{liu2018multiple} as well. Therefore, in this section, we propose a novel framework which can not only deal with non-linearly separable case, but also can learn various weights of different views.

\subsection{Adaptive Weighted Multi-view NMF}
Eq. (\ref{multiview_reg}) utilizes all the data from different views with a consensus $\mG^*$, and treat each view of the same importance. However, in practice, different views may have various weights, which inspires us to propose a formulation that can learn the weights accordingly: 
\begin{equation}\label{multiview_reg_adaptiveWeights}
	\small
\begin{aligned}
     &\min_{\vbeta, \mF, \mG, \mG^*} \sum_{a=1}^v\vbeta_{(a)}^\gamma\{\|\mX^{(a)}-\mF^{(a)}\mG^{(a)} \|_F^2 +  \lambda_a \|\mG^{(a)}-\mG^*\|_F^2\\
    &+ \theta_a\optr{\mG^{(a)}\textbf{L}^{(a)}(\mG^{(a)})^T}\}
    \ \ s.t.  \ \mF^{(a)} ,\mG^{(a)}, \mG^* \geq 0, \sum\vbeta=1,
\end{aligned}
\end{equation}
where $\gamma$ is a hyper-parameter denoting the order of $\vbeta$ which is an integer by default. One can see that when $\gamma=0$, it degenerates into Eq. (\ref{multiview_reg}) where the weight in each view is same. In later section, we will discuss more about its impact on modality selection.
\subsection{Adaptive Weighted Kernel Multi-view NMF}
Most traditional NMF and existing multi-view methods make the clustering/classification analysis based on original data directly, where the accuracy highly relies on the assumption that it is  linearly separable in original space. However, through the extensive experiments on either single view or multi-view analysis, we find the accuracy is rather low as reported in later experiment section. Thus, we propose a \textbf{kernel} version multi-view framework to overcome the difficulty of handling non-linearity via NMF. The basic idea is: instead of clustering on original space, we first map the data into higher dimension $\mPhi: \R^d \rightarrow  \R^p$ (allowing infinite space namely $p=\infty$ such as Gaussian Kernel), and then do clustering based on the mapped sapce.    
To make use \textit{kernel trick}, following the idea of \textit{$K$-means} that each centroid can be represented as a linear combination of data points, we set $\vf^{(a)}=\mPhi(\mX^{(a)})\vp$, and accordingly we have $\mF^{(a)}=\mPhi(\mX^{(a)})\mP$. Also, following the idea of semi-NMF, we remove the nonnegative constraint on $\mP$ such that the learnt features can be more flexible. Finally, we formulate our proposed kernel adaptive multi-view objective as:
\begin{equation}\label{obj_kernel}
	\small
\begin{aligned}
     \sum_{a=1}^v&\vbeta_{a}^\gamma\{\|\mPhi(\mX^{(a)})-\mPhi(\mX^{(a)})\mP^{(a)}\mG^{(a)} \|_F^2 +  \lambda_a \|\mG^{(a)}-\mG^*\|_F^2\\
    + \theta_a&\optr{\mG^{(a)}\textbf{L}^{(a)}(\mG^{(a)})^T}\}
    \ \ s.t. \quad \mG^{(a)}, \mG^* \geq 0, \sum\vbeta=1.
\end{aligned}
\end{equation}

Eq. (\ref{obj_kernel}) not only deals with non-linear data across multi-view but also learns the weight in each view adaptively. 
\section{Optimization}\label{opt}
Given the variables to be optimized, we propose an alternating minimization method to optimize the solution iteratively. 

\textbf{Optimizing $\mP$ in each view}: As there is no constraint on $\mP$, we can simply take the derivative and set it to be $0$ and we have: $\mP\mG\mG^T=\mG^T$, then we have:
\begin{equation}\label{P}
    \mP=\mG^T(\mG\mG^T)^{-1}.
\end{equation}

\textbf{Optimizing $\mG$ in each view}: Since $\mP$ may have mixed signs, multiplicative updating algorithm can't guarantee non-negativity of $\mG$. Therefore, we propose \textit{projected gradient descent} method to make the objective monotonically decrease with update:
\begin{equation}\label{G}
\mG^+=\max\{\mG-\frac{1}{Lips}\nabla_{\mG}\mJ,0\},
\end{equation}
where $Lips$ is the \textit{Lipschitz continuous gradient} and $\mJ$ denotes the objective in Eq. (\ref{obj_kernel}). By definition, we have:  $Lips=2[\sigma_{max}(\mP^T\mK\mP)+\lambda+\theta\sigma_{max}(\mL)], \nabla_{\mG}\mJ=2(\mP^T\mK\mP\mG-\mP^T\mK+\lambda(\mG-\mG^*)+\theta\mG\mL)$, where $\mK(i,j)=\langle \mPhi(\vx_i), \mPhi(\vx_j)\rangle$. With different kernel chosen,  $\langle \mPhi(\vx_i), \mPhi(\vx_j)\rangle$ is different but always computationally economic. For example, if a linear Kernel is chosen, then $\langle \mPhi(\vx_i), \mPhi(\vx_j)\rangle=\vx_i^T\vx_j$; for  polynomial kernel,  $\langle \mPhi(\vx_i), \mPhi(\vx_j)\rangle=(\vx_i^T\vx_j+c)^d$; for Gaussian Kernel $\langle \mPhi(\vx_i), \mPhi(\vx_j)\rangle=exp(-\frac{\|\vx_i - \vx_j \|^2}{2\sigma^2})$, where $c, d$ and $\sigma$ are hyper-parameters.

\textbf{Optimizing $\mG^*$}: By taking the derivative with respect to $\mG^*$ and set it to 0, we have:
\begin{equation}\label{G_star}
    \mG^*=\frac{\sum\vbeta_a^\gamma\lambda_a\mG^{(a)}}{\sum\vbeta_a^\gamma\lambda_a},
\end{equation}
which automatically satisfying the non-negative constraint. 

\textbf{Optimizing $\vbeta$}: For sake of simplicity, we denote the loss in each view as $\vq_a$. Given $\vq$ with other variables fixed in each view, we optimize $\vbeta$ using Lagrangian multipliers:
\begin{equation}\label{lagrangian}
    L(\vbeta,\gamma t)=\sum_{a=1}^v\vbeta_{a}^\gamma\vq_a - \gamma t(\sum_{a=1}^v\vbeta_{a}-1).
\end{equation}
By taking the derivative \wrt $\vbeta$ and $t$, with simple algebra operation we can obtain the optimal solution $\vbeta_a^*=\frac{\vq_a^\frac{1}{1-\gamma}}{\sum_a\vq_a^\frac{1}{1-\gamma}}$. Apparently, $\gamma$ plays a key factor to learn the adaptive weights in each view. Specifically, when $\gamma=1$, only the view with least error will be selected. This can be validated from the solution of $\vbeta$ when $\gamma\rightarrow 1^+$.

Due to space limit, we leave the objective decreasing proof details with update to the supplemental file.
\begin{algorithm}[tb]
\caption{Adaptive weighted kernel multi-view MNF}
\label{alg:algorithm}
\textbf{Input}: 
Multi-view input data $\{\mX^{(1)}, \mX^{(2)}, \cdots, \mX^{(v)}\}$,\\
\textbf{Initialization}:
Feature matrices $\{\mP^{(1)}, \mP^{(2)}, \cdots, \mP^{(v)}\}$, \\
Membership matrices $\{\mG^{(1)}, \mG^{(2)}, \cdots, \mG^{(v)}\}$, \\
Consensus matrix $\mG^*$,
$\vbeta$ (s.t \ $\sum_{a=1}^v\vbeta_{a}=1$), $\lambda$ and $\theta$.\\
\textbf{Output}:$\mG^*$
\begin{algorithmic}[1] 
\STATE Calculate Laplace matrix $\mL^{(a)} = \mD^{(a)} - \mW^{(a)}$.
\REPEAT
\STATE For each view $a$, update $\mP^{(a)}$ as Eq. (\ref{P});
\STATE For each view $a$, update $\mG^{(a)}$ as Eq. (\ref{G});
\STATE Update $\mG^{*}$ as Eq. (\ref{G_star});
\STATE Update $\vbeta$ as Eq. (\ref{lagrangian}).
\UNTIL{converges}
\STATE \textbf{return} $\mG^*$
\end{algorithmic}
\end{algorithm}

\section{Experiments}
In this section we are going to evaluate the performance of the proposed algorithm on ADNI dataset.
\subsection{AD dataset}
The data used to validate our method is obtained from the ADNI database with 4 views: VBM, FreeSurfer, FDG-PET and SNPs with feature dimensions 86, 56, 26 and 1224 respectively. After removing all incomplete samples, we have 88 AD, 174 MCI and 83 HC in total, which will be used for computing the clustering accuracy compared with $\mG^*$. 
\subsection{Baseline methods}
To demonstrate the advantage of our proposed method, we compare it with the following methods including:

\textbf{Single View (SV)}: Vanilla NMF with objective denoted as Eq. (\ref{graph_reg}), in our experiment we run 4 views separately.

\textbf{Feature Concatenation (CNMF)}: A simple and straightforward way to concatenate all features from different views $\mX=\{\mX^{(1)}; \cdots; \mX^{(v)}\}$ and run single view method.

\textbf{(Adaptive Weighted) Multi-view NMF (AW/MNMF)}: which uses original data without kernel mapping as Eq. (\ref{multiview_reg}-\ref{multiview_reg_adaptiveWeights}), it can be regarded as a special case of our proposed method when it is linear kernel.

\subsection{Experiment Setting and Result}
 We conduct grid search with cross-validation to determine the two regularization parameters $\lambda$ and $\theta$, with each being set to be $1$ and $\gamma=2$. We choose different kernels. For polynomial kernel, we set $c=1$ and $d$ varies in $[1,2,3]$. For Gaussian kernel, we set $\sigma$ to be $ exp\{-2,-1,0,\cdots,10\}$.  We found with different settings, our proposed method will converge around  15 iterations, which is very efficient. Table \ref{performance2} shows the clustering results (Accuracy, NMI, Rand and Mirkin's Index) of different methods on the AD dataset. Obviously, our proposed method outperforms the counterparts.
\begin{figure}
	\centering
	\includegraphics[width=.9\linewidth]{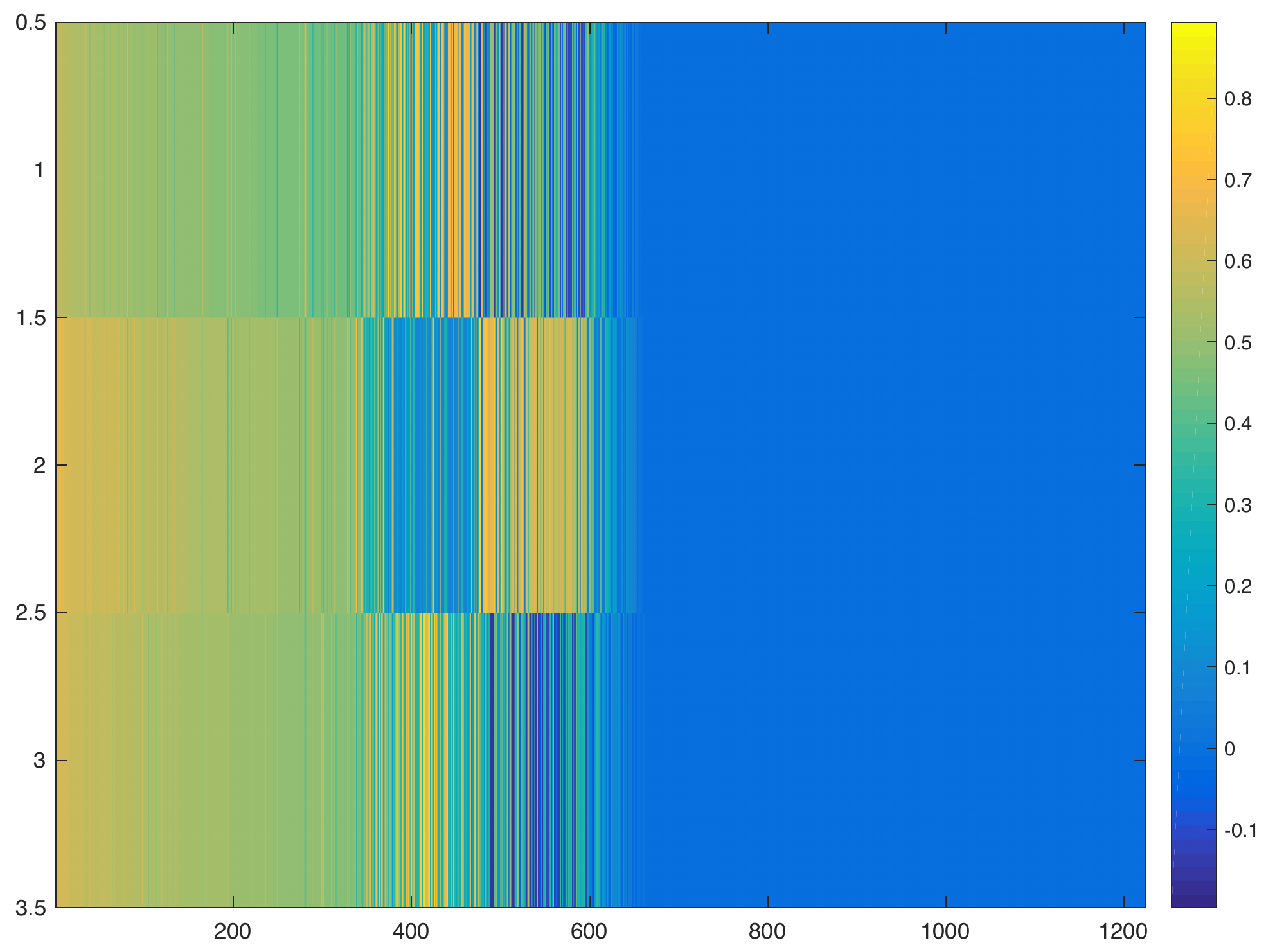}
	\caption{The significance of 1224 SNP to AD (Linear Kernel), with higher value indicates the importance to AD, this could help us narrow the search of SNP causing AD.}
	\label{fig:snp}
\end{figure}
It's interesting to check the feature we learned from our method (for sake of interpretability, we choose linear kernel). The magnitude of $\mF$ represents the importance of a certain feature(SNP) to AD. We plot the heatmap of $\mF$ and sort it by the magnitude as \ref{fig:snp}, it is easy to find that almost 50\% SNP may not be important to AD. For the SNPs in the first half of the fig (which is possibly closely related), we check some with existing clinical discovery. We find that many features given by our prediction do play an important role in AD, for example 
rs3818361, rs10519262 and rs2333227 could be found here. \footnote{
https://www.snpedia.com/index.php/Alzheimer\%27s\_disease}


We also compare the features learned by other views (VBM, FreeSurfer, FDG), following the similar way used in analysis of SNP. We observe that hippocampal measures (LHippocampus,
RHippocampus, LHippVol and RHippVol) are also identified, which
is in accordance with the fact that in the pathological pathway
of AD, medial temporal lobe including hippocampus is firstly
affected, followed by progressive neocortical damage. The thickness measures
of isthmus cingulate (LIsthmCing and RIsthmCing), frontal pole
(LFrontalPole and RFrontalPole) and posterior cingulate gyrus
(LPostCingulate and RPostCingulate) are also selected, which,
again, is accordance with the fact that the GM atrophy of these
regions is high in AD.
\begin{table}
	\begin{center}
		\caption{Average Clustering performance on AD Dataset (\%)}
		\begin{tabular}{*{6}{c}}
			\toprule
			Method & SV& CNMF& MNMF &AWMNMF & \textbf{Ours} \\
			\midrule
			Acc. & $43.25$ & $43.65$& $48.97$ &$50.35$ & $\textbf{54.78}$ \\
			 NMI& $50.55$ & $50.55$&$50.54$ & $50.51$ & $\textbf{54.20}$\\
			  RI& $47.63$ & $47.61$&$40.11$ & $38.09$ & $\textbf{55.87}$\\
			   MI& $52.37$ & $52.39$&$59.89$ & $61.91$ & $\textbf{62.46}$\\
			\bottomrule
		\end{tabular}\label{performance2}
	\end{center}
\end{table}


 \bibliographystyle{IEEEbib}
 \bibliography{refs}
\section{Supplemental}
We now turn to show the details in Section ~\ref{opt}.
\subsection{Optimizing $\mP$ in each view}
While fixing $\vbeta, \mG^*, \mG^{(a)}$, to minimize:
\begin{equation}\label{optO}
	\small
	\begin{aligned}
		\sum_{a=1}^v&\vbeta_{a}^\gamma\{\|\mPhi(\mX^{(a)})-\mPhi(\mX^{(a)})\mP^{(a)}\mG^{(a)} \|_F^2 +  \lambda_a \|\mG^{(a)}-\mG^*\|_F^2\\
		+ \theta&\optr{\mG^{(a)}\textbf{L}^{(a)}(\mG^{(a)})^T}\}
		\ \ s.t. \quad \mG^{(a)}, \mG^* \geq 0, \sum\vbeta=1,
	\end{aligned}
\end{equation}
is equivalent to minimize:
\begin{equation}
	\|\mPhi(\mX^{(a)})-\mPhi(\mX^{(a)})\mP^{(a)}\mG^{(a)} \|_F^2.
\end{equation}
For sake of simplicity, we remove the superscript $(a)$ and we have:
\begin{equation}
	\begin{aligned}
	\mJ&=\|\mPhi(\mX)-\mPhi(\mX)\mP\mG \|_F^2\\
	&=\optrb{(\mPhi(\mX)-\mPhi(\mX)\mP\mG)^T(\mPhi(\mX)-\mPhi(\mX)\mP\mG)}\\
	&=\optrb{\mK-2\mK\mP\mG+\mG^T\mP^T\mK\mP\mG},
\end{aligned}
\end{equation}
where by definition $\mK=\mPhi(\mX)^T\mPhi(\mX)$ which is symmetric. Taking the derivative \wrt $\mP$ and set it to be 0, we have:
\begin{equation}
	\begin{aligned}
	\nabla_{\mP} \mJ=2(&\mK\mP\mG\mG^T-\mK\mG^T)=0\\
	\implies &\mK\mP\mG\mG^T=\mK\mG^T\\
	\implies &\mP=\mG^T(\mG\mG^T)^{-1}
\end{aligned}
\end{equation}
\subsection{Optimizing $\mG$ in each view}
Following the way we optimize $\mP$, to minimize $\mG$ in each view, it is equivalent to minimize:
\begin{equation}\label{optG}
	\|\mPhi(\mX)-\mPhi(\mX)\mP\mG \|_F^2+\lambda\|\mG-\mG^*\|_F^2+\theta \optrb{\mG\mL\mG^T}
\end{equation}
with $\mG\geq0$. In the following, we are to show that if $\mG^+=max\{\mG-t*\nabla_{\mG}\mJ,0\}$, where $t=\frac{1}{Lips}$, then the objective is monotonically decreasing. We recognize this method is simply Projected Gradient Desent and refer the reviewer to relative references as the proof to objective decreasing part is standard and we omit the details here. To begin with, we are going to determine \textit{Lipschitz continuous constant}. 
\begin{theorem}
A differentiable function $\mJ$ is said to have an \textit{L-Lipschitz continuous gradient} if for some $L>0$ 
\begin{equation}
\lVert \nabla \mJ(\mG^+) - \nabla \mJ(\mG)\rVert \le L \lVert \mG^+-\mG\rVert,~\forall \mG^+,\mG.
\end{equation}
\end{theorem}
According to Eq. (\ref{optG}), we have $\nabla\mJ(\mG)=2(\mP^T\mK\mP\mG-\mP^T\mK+\lambda(\mG-\mG^*)+\theta\mG\mL)$, therefore:
\begin{equation}
	\begin{aligned}
		&\lVert \nabla \mJ(\mG^+) - \nabla \mJ(\mG)\rVert\\=&2\lVert \mP^T\mK\mP(\mG^+-\mG)+\lambda(\mG^+-\mG)+\theta (\mG^+-\mG)\mL \rVert\\
		\le&2\lVert \mP^T\mK\mP(\mG^+-\mG)\|+\|\lambda(\mG^+-\mG)\|+\|\theta (\mG^+-\mG)\mL \rVert\\
		\le&2(\lVert \mP^T\mK\mP\rVert+\lambda+\theta\lVert\mL\rVert)\lVert\mG^+-\mG\rVert\\
		=&L\lVert\mG^+-\mG\rVert
	\end{aligned}
\end{equation}
where $L=2[\sigma_{max}(\mP^T\mK\mP)+\lambda+\theta\sigma_{max}(\mL)]$. The second and third lines follow from Subadditivity and Submultiplicative  Inequality ($\|A+B\|\le\|A\|+\|B\|, \|AB\|\le\|A\|\|B\|$ respectively), therefore $G^+=max\{\mG-\frac{1}{L}\nabla_{\mG}\mJ,0\}$ will make the objective monotonically non-increasing.
\subsection{Optimizing $\mG^*$}
Taking the derivative of Eq. (\ref{optO}) \wrt $\mG^*$ and set it to be 0, we have:
\begin{equation}
	2\sum_{a}\vbeta_{a}^\gamma\lambda_a(\mG^*-\mG^{(a)})=0,
\end{equation}
with simple reformulation, we have
\begin{equation}
\mG^*=\frac{\sum\vbeta_a^\gamma\lambda_a\mG^{(a)}}{\sum\vbeta_a^\gamma\lambda_a}.
\end{equation} 
Given the nonnegative of $\mG^{(a)}, \lambda$ and $\vbeta$, $\mG^*$ automatically satisfies the non=negative constraint.
\subsection{Optimizing $\vbeta$}
While fixing all the rest variables and denote the objective in each view as $\vq_a$, we utilize Lagrangian Multiplier and formulate the objective as:
\begin{equation}\label{lag}
	L(\vbeta,\gamma t)=\sum_{a=1}^v\vbeta_{a}^\gamma\vq_a - \gamma t(\sum_{a=1}^v\vbeta_{a}-1).
\end{equation}
By taking the derivative \wrt $\vbeta_a$ ($1\le a\le v$), we have:
\begin{equation}
	\gamma\vq_a\vbeta_a^{\gamma-1}-\gamma t=0
\end{equation}
which implies 
\begin{equation}\label{beta}
\vbeta_a=(\frac{t}{\vq_a})^{\frac{1}{\gamma-1}}
\end{equation}
As $\sum_{a}\vbeta_{a}=1$, we  have: 
\begin{equation}
t=(\sum\vq_a^{\frac{1}{1-\gamma}})^{1-\gamma}
\end{equation}
Now pluggin back to Eq. (\ref{beta}), we have:
\begin{equation}
 \vbeta_a=\frac{\vq_a^\frac{1}{1-\gamma}}{\sum_a\vq_a^\frac{1}{1-\gamma}}
\end{equation}
which is the weight of each view.
\end{document}